\documentclass[journal,draftclsnofoot,letterpaper,10pt,onecolumn]{IEEEtran}
\usepackage[cmex10]{amsmath}
\usepackage{graphicx}    
\newtheorem{thm}{Theorem}

\newtheorem{axm}{Axiom}
\newtheorem{rmk}{Remark}
\newtheorem{defi}{Definition}

\ifCLASSINFOpdf
\else
\fi
\begin{document}
%
\title{Analysis and Extension of the Evidential Reasoning Algorithm for Multiple Attribute Decision Analysis with Uncertainty}
%
%
%

\author{Lianmeng~Jiao and Xiaojiao Geng
\thanks{This work was funded by the National Natural Science Foundation of China (Grant Nos. 61790552 and 61801386), the Natural Science Basic Research Plan in Shaanxi Province of China (Grant No. 2018JQ6043) and the Aerospace Science and Technology Foundation of China.}
\thanks{L. Jiao and X. Geng are with the School of Automation, Northwestern Polytechnical University, Xi'an 710072, P. R. China (e-mail: jiaolianmeng@nwpu.edu.cn, gengxiaojiao@mail.nwpu.edu.cn).}
}

\maketitle

\begin{abstract}
In multiple attribute decision analysis (MADA) problems, one often needs to deal with assessment information with uncertainty. The evidential reasoning approach is one of the most effective methods to deal with such MADA problems. As kernel of the evidential reasoning approach, an original evidential reasoning (ER) algorithm was firstly proposed by Yang et al, and later they modified the ER algorithm in order to satisfy the proposed four synthesis axioms with which a rational aggregation process needs to satisfy. However, up to present, the essential difference of the two ER algorithms as well as the rationality of the synthesis axioms are still unclear. In this paper, we analyze the ER algorithms in Dempster-Shafer theory (DST) framework and prove that the original ER algorithm follows the reliability discounting and combination scheme, while the modified one follows the importance discounting and combination scheme. Further we reveal that the four synthesis axioms are not valid criteria to check the rationality of one attribute aggregation algorithm. Based on these new findings, an extended ER (E$^2$R) algorithm is proposed to take into account both the reliability and importance of different attributes, which provides a more general attribute aggregation scheme for MADA with uncertainty. A motorcycle performance assessment problem is examined to illustrate the proposed algorithm.
\end{abstract}

\begin{IEEEkeywords}
Multiple attribute decision analysis (MADA), Dempster-Shafer theory (DST), evidential reasoning (ER) algorithm, reliability, importance.
\end{IEEEkeywords}

%
\IEEEpeerreviewmaketitle

\section{Introduction}
%
%
%
%
In many real-world applications, multiple attribute decision analysis (MADA) problems are characterized using both quantitative and qualitative attributes \cite{Hwang81}. Usually, the qualitative attributes could only be properly assessed with subjective human judgments \cite{Perez14} and hence are inevitably associated with uncertainties because of human being's inability to provide complete judgments (probabilistic uncertainty) \cite{Taha14} or vagueness of the meanings of assessments (fuzzy uncertainty) \cite{Yang06,Gupta16,Zhang16}. In MADA society, multiple attribute utility theory (MAUT) \cite{Moore79,Belton02} and analytical hierarchy process (AHP) \cite{Saaty80,Ishizaka11,Pedrycz11} are the most popular methods, in which an alternative is assessed on each attribute by either a single number or an interval value. However, in many decision situations, using a single number or interval to represent a judgment proves to be difficult and may be unacceptable \cite{Guo09}. Information may be lost or distorted in the process of aggregating different types of uncertain information, such as probability distributions or fuzzy assessments.

During the last two decades, an evidential reasoning approach has been proposed and developed for MADA with uncertainty \cite{Yang94a,Yang94b,Yang01a,Yang02a,Yang02b}. Different from the traditional MADA methods, a distributed evaluation analysis model is used to characterize the assessments for both quantitative and qualitative attributes. In this model, each attribute is assessed using a set of collectively exhaustive and mutually exclusive assessment grades, and so probabilistic uncertainty can be characterized by a belief structure and fuzzy uncertainty by linguistic variables. Due to its well performance for addressing uncertainty, the evidential reasoning approach has been applied to many MADA problems, such as engineering design \cite{Martinez07,Chin09,Browne13}, safety and risk assessment \cite{Wang01,Hu10,Zhou16}, supply chain management \cite{Sonmez02,Ren09}, environment management \cite{Wang06,Yao10}, etc.

The kernel of the evidential reasoning approach is an evidential reasoning (ER) algorithm developed on the basis of the evaluation analysis model. An original ER algorithm was firstly proposed by Yang and Singh \cite{Yang94a}. Later, in order to develop theoretically sound methods and tools for dealing with uncertain MADA problems, Yang and Xu \cite{Yang02b} proposed four synthesis axioms with which a rational aggregation process needs to satisfy. They also showed that the original ER algorithm failed to satisfy all these axioms, and exactly guided by the aim, the authors proposed a modified ER algorithm. Consequently, for later applications of MADA with uncertainty, the modified ER algorithm was widely used and the original one was just abandoned. However, the problem is whether the original ER algorithm is wrong, or more generally, whether only the algorithms satisfying the four axioms are rational \cite{Xu12}. It's important to discover the essential difference of the two ER algorithms and the rationality of the synthesis axioms so that we can take a right choice when using these algorithms for resolving MADA problems. Recently, several works have been done intending to analyze the ER algorithms as well as the four synthesis axioms. Huynh et al. \cite{Huynh06} analyzed the ER algorithms explicitly in terms of Dempster-Shafer theory (DST) \cite{Dempster67,Shafer76} and in the spirit of such an analysis, two other aggregation algorithms which also satisfy the synthesis axioms were given. Although much work has been done to analyze the formulation of the original and modified ER algorithms, they have not discovered the nature of the two algorithms. More recently, Durbach \cite{Durbach12} conducted an empirical test of the four synthesis axioms and showed that evaluations which invoke the axioms frequently violate them, but the essential reason is still not clear.

As we know that, the two ER algorithms are both developed on the basis of the evaluation analysis model and the combination rule of DST. The major difference is how to address the weights of different attributes. New findings about reliability and importance of evidence in DST \cite{Smarandache10,Jiao16} provide us a new way to reanalyze the ER algorithms for MADA with uncertainty. The reliability can be seen as an objective property of a source of evidence representing its capability to provide correct assessment, while the importance is a subjective property of a source of evidence reflecting the decision maker's subjective preference. Accordingly, they should be addressed with different schemes, namely reliability discounting $\&$ combination scheme and importance discounting $\&$ combination scheme, respectively. For the MADA problems, we believe that the weights of attributes can be interpreted as the evaluation reliability or the relative importance of attributes, whereas in the original and modified ER algorithms, the two properties make no difference. In this paper, we analyze the ER algorithms in terms of reliability and importance of evidence in DST and extend them to a more general scheme. The contribution of this paper is threefold:
\begin{enumerate}
  \item We prove that the original ER algorithm follows the reliability discounting $\&$ combination scheme, while the modified ER algorithm follows the importance discounting $\&$ combination scheme.
  \item We explain the synthesis axioms from the aspects of reliability and importance of evidence and reveal that the four synthesis axioms are not valid criteria to check the rationality of one attribute aggregation algorithm.
  \item We introduce an extension of the ER algorithm, called E$^2$R algorithm, which provides a more general attribute aggregation scheme for MADA with uncertainty.
\end{enumerate}

The rest of this paper is organized as follows. The basic notions in MADA problem with uncertainty and the evidential reasoning approach are recalled in Section \ref{sec2}. The reliability and importance of evidence in DST useful for later analysis are given in Section \ref{sec3}. Section \ref{sec4} gives a deep analysis of the ER algorithms as well as the synthesis axioms in DST framework. A more general E$^2$R algorithm is proposed in Section \ref{sec5}, and then a motorcycle performance assessment problem is examined in Section \ref{sec6}. Finally, Section \ref{sec7} presents some concluding marks.

\section{Background} \label{sec2}
\subsection{MADA Problem with Uncertainty} \label{sec2A}
\begin{figure}[!ht]
\centering
\includegraphics[scale=0.7]{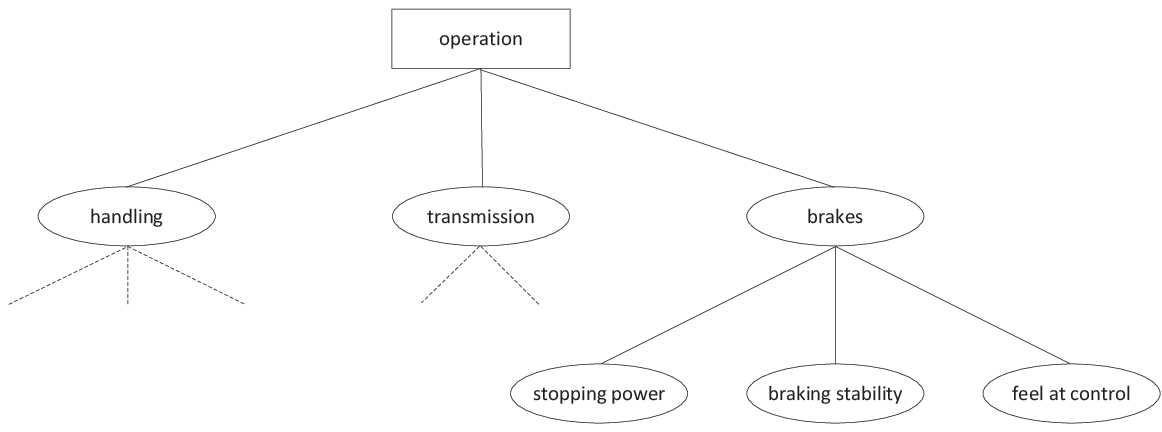}
\caption{Evaluation hierarchy for \emph{operation} of a motorcycle \cite{Yang02b}}
\label{fig1}
\end{figure}

We describe the MADA problem with uncertainty considering a motorcycle performance assessment example taken from \cite{Yang02b}. The hierarchy for evaluating the \emph{operation} of a motorcycle is illustrated in Fig.\ref{fig1}.

To subjectively evaluate the attributes of each decision alternative, a set of evaluation grades are given as
\[\mathcal{H} = \left\{ {{H_1}, \cdots ,{H_n}, \cdots ,{H_N}} \right\},\]
where ${H_n} (n=1,\cdots,N)$ are called evaluation grades to which the state of an attribute can be evaluated. For example, for evaluating the quality of the \emph{operation} of a motorcycle, the set of evaluation grades can be defined as
\begin{equation}
\mathcal{H} = \left\{ {poor({H_1}),indifferent({H_2}),average({H_3}),good({H_4}),excellent({H_5})} \right\}. \label{eq_evaluation_grade}
\end{equation}

In subjective attribute evaluation process, judgments may sometimes be uncertain. For example, in evaluating the \emph{brakes} of a motorcycle, expert's judgments may be
\begin{itemize}
  \item Its \emph{stopping power} is \emph{excellent} with a confidence degree of $1$.
  \item Its \emph{braking stability} is \emph{average} with a confidence degree of $0.4$ and is \emph{good} with a confidence degree of $0.6$.
  \item Its \emph{feel at control} is evaluated to be \emph{good} with a confidence degree of $0.5$ and to be \emph{excellent} with a confidence degree of $0.3$.
\end{itemize}
In the above three statements, expert's judgment for \emph{stopping power} is exact and certain; the judgment for \emph{braking stability} is probabilistic; and the judgment for \emph{feel at control} is incomplete (the total confidence degree is smaller than $1$). The evidential reasoning approach \cite{Yang94a,Yang94b,Yang02b} developed based on Dempster-Shafer theory has provided a powerful tool for dealing with such uncertain judgments aggregation problem. The kernel of the evidential reasoning approach is evidential reasoning (ER) algorithm developed on the basis of an evaluation analysis model. In the following, after a brief description of the utilized evaluation analysis model, we give a summary of the existing ER algorithms.

\subsection{Evaluation Analysis Model} \label{sec2B}
\begin{figure}[!ht]
\centering
\includegraphics[scale=0.8]{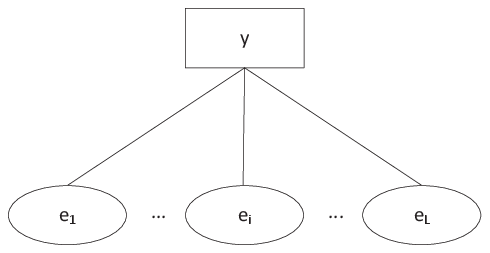}
\caption{Two-level evaluation hierarchy}
\label{fig2}
\end{figure}

Yang and Singh \cite{Yang94a} proposed an evaluation analysis model to represent those uncertain subjective judgments specified in the preceding subsection. Let us consider a simple two-level evaluation hierarchy with a general attribute denoted by $y$ at the top level and $L$ basic attributes denoted by $e_i$ ($i=i,\cdots, L$) at the bottom level (as shown in Fig.\ref{fig2}). Let $E = \left\{ e_1, \cdots, e_i, \cdots, e_L \right\}$ be the set of basic attributes and $W = \left\{ w_1, \cdots, w_i, \cdots, w_L \right\}$ be the set of corresponding attribute weights.

Given the set of evaluation grades $\mathcal{H} = \left\{ H_1, \cdots ,H_n, \cdots ,H_N \right\}$ designed as distinct standards for assessing an attribute, then an assessment for ${e_i}$ of a decision alternative can be mathematically represented in terms of the distribution \cite{Yang02b}
\begin{equation}
S(e_i) = \left\{ (H_n,\beta_{n,i})|n = 1, \cdots ,N \right\},\text{~for~} i = 1, \cdots L, \label{eq_dist}
\end{equation}
where $\beta_{n,i}$ is the belief degree supporting $H_n$. Note that the sum of belief degrees is not always equal to 1 (e.g. the third assessment in the preceding subsection).

Denote $\beta_n$ the belief degree to which the general attribute $y$ is assessed to the evaluation grade of $H_n$. The problem now is to compute $\beta_n$ ($n = 1, \cdots ,N$), by combining the assessments from all associated basic attributes $e_i$ ($i = 1, \cdots ,L$) as given in Eq.(\ref{eq_dist}). The following ER algorithms were designed for this purpose.

\subsection{Summary of the ER Algorithms} \label{sec2C}
An original ER algorithm was proposed in \cite{Yang94a} aggregating the assessments of basic attributes given in Eq.(\ref{eq_dist}) to obtain the assessment of the general attribute.

\textbf{Original ER algorithm:}
\begin{enumerate}
  \item \emph{Weighting the belief distribution}. Denote $m_{n,i}$ the basic probability mass that the basic attribute ${e_i}$ supports the hypothesis that the attribute $y$ is assessed to the evaluation grade $H_n$, and $m_{\mathcal{H},i}$ the remaining probability mass unassigned to any individual grade. These quantities are calculated by weighting the belief distribution given in Eq.(\ref{eq_dist}) as
\begin{equation}
\begin{array}{*2{l}}
{m_{n,i}} &= {w_i}{\beta_{n,i}},\text{~~for~} n = 1, \cdots ,N \\
{m_{\mathcal{H},i}} &= 1 - \sum\nolimits_{n = 1}^N {m_{n,i}}  = 1 - {w_i}\sum\nolimits_{n = 1}^N {\beta_{n,i}}.
\end{array} \label{eq_OER_w}
\end{equation}

  \item \emph{Aggregation process}. Denote $E_{I(i)} = \{e_1, \cdots, e_i\}$ the subset of first $i$ basic attributes. Then the combination results $m_{n,I(i+i)}$ and $m_{\mathcal{H},I(i+1)}$ for the first $i+1$ assessments can be recursively calculated as
\begin{equation}
\begin{array}{*3{l}}
\{H_n\} : &{m_{n,I(i + 1)}} &= {K_{I(i + 1)}}\left[{m_{n,I(i)}}{m_{n,i + 1}} + {m_{n,I(i)}}{m_{\mathcal{H},i + 1}} + {m_{\mathcal{H},I(i)}}{m_{n,i + 1}}\right], \text{~for~} n = 1,2, \cdots N\\
\mathcal{H} : &{m_{\mathcal{H},I(i + 1)}} &= {K_{I(i + 1)}}\left[{m_{\mathcal{H},I(i)}}{m_{\mathcal{H},i + 1}}\right]\\
&{K_{I(i + 1)}} &= {\left[1 - \sum\limits_{j = 1}^N {\sum\limits_{p = 1; p \ne j}^N {{m_{j,I(i)}}{m_{p,i + 1}}}} \right]^{ - 1}}, \text{~for~} i = 1,2, \cdots ,L - 1,
\end{array} \label{eq_OER_agr}
\end{equation}
where, ${K_{I(i + 1)}}$ is a normalizing factor so that $\sum\nolimits_{n = 1}^N {{m_{n,I(i + 1)}}}  + {m_{\mathcal{H},I(i + 1)}} = 1$.

  \item \emph{Generation of combined belief degree}. After aggregating the assessments for all the $L$ basic attributes, the combined belief degree for the general attribute $y$ are given by
\begin{equation}
\begin{array}{*3{l}}
\{H_n\} : &{\beta_n} &= m_{n,I(L)}, \text{~for~} n = 1,2, \cdots N\\
\mathcal{H} : &{\beta_\mathcal{H}} &= m_{\mathcal{H},I(L)}.
\end{array} \label{eq_OER_norm}
\end{equation}
\end{enumerate}

Intending to develop theoretically sound methods for dealing with uncertain MADA problems, Yang and Xu \cite{Yang02b} later proposed four synthesis axioms based on the evaluation analysis model and they believe that a rational aggregation process needs to satisfy these synthesis axioms.
\begin{axm}[Independence]
If no basic attribute is assessed to an evaluation grade at all, then the general attribute should not be assessed to the same grade either.
\end{axm}

\begin{axm}[Consensus]
If all basic attributes are precisely assessed to an individual grade, then the general attribute should also be precisely assessed to the same grade.
\end{axm}

\begin{axm}[Completeness]
If all basic attributes are completely assessed to a subset of grades, then the general attribute should be completely assessed to the same subset as well.
\end{axm}

\begin{axm}[Incompleteness]
If any basic assessment is incomplete, then a general assessment obtained by aggregating the incomplete and complete
basic assessments should also be incomplete.
\end{axm}

As demonstrated in \cite{Yang02b}, the original ER algorithm only satisfies the \emph{independence} axiom. Under such a consideration, a modified ER algorithm that satisfies all these synthesis axiom precisely is developed in \cite{Yang02b}.

\textbf{Modified ER algorithm:}
\begin{enumerate}
  \item \emph{Weight normalization}. In the modified ER algorithm, the weights ${w_i}$ ($i = 1, \cdots ,L$) of basic attributes are normalized such that $0 \le {w_i} \le 1,$ and
\begin{equation}
\sum\limits_{i = 1}^L {w_i}  = 1. \label{eq_MER_w}
\end{equation}

  \item \emph{Remaining probability mass decomposition}. The remaining probability mass ${m_{\mathcal{H},i}}$ given in Eq.(\ref{eq_OER_w}) is decomposed into two parts: ${m_{\mathcal{H},i}} = {\bar m_{\mathcal{H},i}} + {\tilde m_{\mathcal{H},i}}$, where
\begin{equation}
\bar m_{\mathcal{H},i} = 1 - w_i \text{~and~} \tilde m_{\mathcal{H},i} = {w_i}\left(1 - \sum\nolimits_{n = 1}^N {\beta_{n,i}}\right). \label{eq_MER_decom}
\end{equation}
$\bar m_{\mathcal{H},i}$ represents the degree to which other attributes can play a role in the assessment, while ${\tilde m_{\mathcal{H},i}}$ is caused due to the incompleteness in the assessment.

  \item \emph{Aggregation process with decomposed masses}. The combination process with the two decomposed masses ${\bar m_{\mathcal{H},i}}$ and ${\tilde m_{\mathcal{H},i}}$ can be recursively carried out as
\begin{equation}
\begin{array}{*2{l}}
\{H_n\} : m_{n,I(i + 1)} &= K_{I(i + 1)}\left[{m_{n,I(i)}}{m_{n,i + 1}} + {m_{n,I(i)}}({{\bar m}_{\mathcal{H},i + 1}} + {{\tilde m}_{\mathcal{H},i + 1}}) + ({{\bar m}_{\mathcal{H},I(i)}} + {{\tilde m}_{\mathcal{H},I(i)}}){m_{n,i + 1}}\right],\\
&\text{~~~for~} n = 1,2, \cdots N\\
\mathcal{H} : {{\tilde m}_{\mathcal{H},I(i + 1)}} &= K_{I(i + 1)}\left[{{\tilde m}_{\mathcal{H},I(i)}}{{\tilde m}_{\mathcal{H},i + 1}} + {{\tilde m}_{\mathcal{H},I(i)}}{{\bar m}_{\mathcal{H},i + 1}} + {{\bar m}_{\mathcal{H},I(i)}}{{\tilde m}_{\mathcal{H},i + 1}}\right]\\
\mathcal{H} : {{\bar m}_{\mathcal{H},I(i + 1)}} &= K_{I(i + 1)}\left[{{\bar m}_{\mathcal{H},I(i)}}{{\bar m}_{\mathcal{H},i + 1}}\right],
\end{array} \label{eq_MER_agr}
\end{equation}
where $K_{I(i + 1)}$ is the same as in Eq.(\ref{eq_OER_agr}).

  \item \emph{Belief degree normalization}. After all $L$ assessments have been aggregated, the combined belief degree are generated by assigning ${\bar m_{\mathcal{H},I(L)}}$ back to all individual grades proportionally
\begin{equation}
\begin{array}{*3{l}}
\{H_n\} :& \beta_n &= \frac{{m_{n,I(L)}}}{1 - {{\bar m}_{\mathcal{H},I(L)}}}, \text{~for~} n = 1,2, \cdots N\\
\mathcal{H} : &\beta_\mathcal{H} &= \frac{{{\tilde m}_{\mathcal{H},I(L)}}}{1 - {{\bar m}_{\mathcal{H},I(L)}}}.
\end{array} \label{eq_MER_norm}
\end{equation}
\end{enumerate}

\section{Reliability and Importance of Evidence in DST} \label{sec3}
\subsection{Basic Concepts in DST}
In DST \cite{Shafer76}, a problem domain is represented by a finite set $\Theta  = \{\theta_1, \theta_2, \cdots, \theta_n\}$ of mutually exclusive and exhaustive hypotheses called the \emph{frame of discernment}. A \emph{basic belief assignment} (BBA) expressing the belief committed to the elements of ${2^\Theta }$ by a given source of evidence is a mapping function $m( \cdot)$: ${2^\Theta} \to [0,1]$, such that
\begin{equation}
m(\emptyset) = 0 \text{~and~} \sum\limits_{A \in {2^\Theta}} {m(A)}  = 1.
\end{equation}
Elements $A \in {2^\Theta }$ having $m(A) > 0$ are called \emph{focal elements} of the BBA $m(\cdot)$. A BBA $m(A)$ measures the degree of belief exactly assigned to a proposition $A$. The mass assigned to $\Theta $, i.e. $m(\Theta )$, is referred to as the degree of \emph{global ignorance}.

Shafer \cite{Shafer76} also defines the \emph{belief function} and \emph{plausibility function} of $A \in {2^\Theta }$ as follows
\begin{equation}
Bel(A) = \sum\limits_{B \subseteq A} {m(B)} \text{~and~} Pl(A) = \sum\limits_{B \cap A \ne \emptyset } {m(B)}.
\end{equation}
$Bel(A)$ represents the exact support to $A$ and its subsets, and $Pl(A)$ represents all possible support to $A$ and its subsets. The interval $[Bel(A),Pl(A)]$ can be seen as the lower and upper bounds of support to $A$.

For decision-making support, the \emph{pignistic probability} $BetP (A)$ \cite{Smets94} is commonly used to approximate the unknown probability in $[{\text {Bel}} (A),{\text {Pl}} (A)]$, given by

\begin{equation}
{\text {BetP}} (A) = \sum\limits_{B \subseteq \Theta; A \cap B \ne \emptyset} {\frac{{\left| {A \cap B} \right|}}{{\left| B \right|}}} {\text m} (B),  \label{eq_BetP}
\end{equation}
where, $\left| X \right|$ stands for the cardinality of the set $X$.

\subsection{Reliability Discounting $\&$ Combination Scheme} \label{sec3B}
Within the framework of DST, the reliability of evidence is widely considered in evidence aggregation process due to sources' limitation to provide totally accurate information \cite{Mercier08,Elouedi04}. As in \cite{Smarandache10}, the definition of reliability for a source of evidence in the context of DST is given as follows.
\begin{defi}[Reliability of a source of evidence]
The reliability, as an objective property of a source of evidence, represents its capability to provide correct measure or assessment of the considered problem.
\end{defi}

The reliability of a source of evidence is generally considered according to Shafer's discounting method \cite{Shafer76}. Mathematically, Shafer's discounting operation (detonated as $ \otimes $) for taking into account the reliability factor $\alpha  \in [0,1]$ of a given source with a BBA  ${m} ( \cdot )$ and a frame $\Theta $ is defined as
\begin{equation}
\alpha  \otimes {m}(A) \buildrel \Delta \over = {{{m}}^\alpha }(A) = \left\{ {\begin{array}{*{20}{l}}
   \alpha {{m}}(A),&{\text{for~}}A \in {2^\Theta }{\text{~ and ~}} A \ne \Theta   \\
   \alpha {{m}}(\Theta ) + (1 - \alpha ),&{\text{for~}} A = \Theta.  \\
\end{array}} \right.\label{eq_RD}
\end{equation}
As shown in Eq.(\ref{eq_RD}), Shafer's discounting method multiplies the masses of focal elements by the reliability factor $\alpha$, and transfers all the remaining discounted mass $1 - \alpha$ to the global ignorance set $\Theta $.

After the reliability discounting operation, Dempster's rule of combination \cite{Shafer76} (the orthogonal sum operation $ \oplus $) is used to combine two discounted pieces of evidence represented by two BBAs ${m} _1^{{\alpha _1}}( \cdot )$ and ${m} _2^{{\alpha _2}}( \cdot )$ to generate a new BBA
\begin{equation}
{m} _1^{\alpha_1} \oplus {m} _2^{\alpha_2} ( A ) \buildrel \Delta \over = {m} _{12}^{{D}}(A)
= \left\{ {\begin{array}{*{20}{l}}
   0, &{\text{~for~}}A = \emptyset  \\
   \frac{{\sum\limits_{B,C \in {2^\Theta };B \cap C = A} {{m} _1^{{\alpha _1}}(B){m} _2^{{\alpha _2}}(C)} }}{{1 - \sum\limits_{B,C \in {2^\Theta };B \cap C = \emptyset } {{m} _1^{{\alpha _1}}(B){m} _2^{{\alpha _2}}(C)} }}, &{\text{~for~}}A \in {2^\Theta }{\text{~and~}}A \ne \emptyset. \\
\end{array}} \right.\label{eq_RC}
\end{equation}
As described in \cite{Shafer76}, Dempster's rule of combination is both commutative and associative.

\subsection{Importance Discounting $\&$ Combination Scheme} \label{sec3C}
Different from the concept of reliability, the definition of importance for a source of evidence in the context of DST is given as follows \cite{Jiao16}.
\begin{defi}[Importance of a source of evidence]
The importance, as a subjective property of a source of evidence, represents the weight of that source which reflects the decision maker's subjective preference in the fusion process.
\end{defi}

\begin{rmk} \label{rek_ID}
The importance of a source of evidence can be characterized by an importance factor, denoted $\beta \in [0,1]$. Different from the reliability defined previously, the defined importance is a relative concept. In other words, the importance of one source is meaningless without comparing with others. Hence, the importance factors $\beta _i \in [0,1]$ should satisfy the normalization constraint $\sum\nolimits_{i = 1}^N {{\beta _i}}  = 1$, where $N$ is the number of sources of evidence involved in the fusion process.
\end{rmk}

To address the importance of sources of evidence, an importance discounting operation as the counterpart of Shafer's discounting operation is developed in \cite{Jiao16}. Mathematically, the importance discounting operation (detonated as $ \odot $) of a source of evidence having the importance factor $\beta \in [0,1]$ and associated BBA ${m} ( \cdot )$ is defined as
\begin{equation}
\beta  \odot {m} (A) \buildrel \Delta \over = {{m} ^\beta }(A) = \left\{ {\begin{array}{*{20}{l}}
   \beta {m} (A), &{\text{~for~}}A \in {2^\Theta } \\
   1 - \beta, &{\text{~for~}}A = \Omega \\
\end{array}} \right. \label{eq_ID}
\end{equation}
where, $\Omega$ is the power set of $\Theta$, i.e. $\Omega=2^\Theta$, which characterizes the indecisiveness among the subset of ${2^\Theta }$.

It is worth noting that, after the importance discounting operation, the original BBA is extended with new belief assignment ${{m}^\beta }(\Omega )$. So the classical Dempster's rule of combination needs to be extended to combine different importance BBAs (IBBAs, for short) \cite{Jiao16}. Mathematically, with two pieces of evidence having importance factors ${\beta _1}$ and ${\beta _2}$ represented by two IBBAs ${m} _1^{{\beta _1}}( \cdot )$ and ${m} _2^{{\beta _2}}( \cdot )$ defined on ${2^\Theta }\cup\{\Omega\}   \buildrel \Delta \over = {2^{{\Theta ^ + }}}$, the extended Dempster's rule of combination is defined as
\begin{equation}
{m} _1^{{\beta _1}} \oplus {m} _2^{{\beta _2}}( A ) \buildrel \Delta \over = {m} _{12}^{{{ED}}}(A)
= \left\{ {\begin{array}{*{20}{l}}
   0, &{\text{for~}}A = \emptyset  \\
   \frac{{\sum\limits_{B,C \in {2^{{\Theta ^ + }}};B \cap C = A} {{m} _1^{{\beta _1}}(B){m} _2^{{\beta _2}}(C)} }}{{1 - \sum\limits_{B,C \in {2^{{\Theta ^ + }}};B \cap C = \emptyset } {{m} _1^{{\beta _1}}(B){m} _2^{{\beta _2}}(C)} }}, &{\text{for~}}A \in {2^{{\Theta ^ + }}}{\text{~and~}}A \ne \emptyset.  \\
\end{array}} \right. \label{eq_IC}
\end{equation}
Similar with the Dempster's rule of combination, the extended Dempster's rule of combination is also based on the orthogonal sum operation $\oplus$, so it is also both commutative and associative.

Since we usually need to work with normal BBA for decision-making support, after combining all the importance discounted IBBAs using the extended Dempster's rule of combination, the fusion result ${m}^{{ED}}(\cdot)$ will be normalized by redistributing the mass of belief committed to the set $\Omega$ to the other focal elements proportionally to their masses as follows \cite{Jiao16}
\begin{equation}
{{m} ^{{{norm}}}}(A) = \frac{{{{m} ^{{{ED}}}}(A)}}{{1 - {{m} ^{{{ED}}}}(\Omega )}},{\text{~}} \forall A \in {2^\Theta }. \label{eq_Norm}
\end{equation}

\section{Analysis of the ER Algorithms in DST Framework} \label{sec4}
The new development about reliability and importance of evidence in DST described in the preceding section provides us a new way to analyze the ER algorithms for MADA with uncertainty. In this section, we first represent the uncertain MADA problem in DST framework. Then the original and modified ER algorithms are analyzed in terms of reliability and importance of evidence, respectively. Finally, with the new findings about the potential nature of the two ER algorithms, the rationality of the synthesis axioms is discussed in a more reasonable way.

\subsection{Problem Representation in DST Framework}
Let us reconsider the available information given for a MADA problem with uncertainty in the two-level evaluation hierarchy, as shown in Fig.\ref{fig2}:
\begin{enumerate}
  \item the uncertain assessments $S(e_i) = \left\{ (H_n,\beta_{n,i})|n = 1, \cdots, N \right\}$ of basic attributes ${e_i}$, for $i = 1, \cdots, L$;
  \item the weights ${w_i}$ of basic attributes ${e_i}$, for $i = 1, \cdots, L$.
\end{enumerate}

In terms of DST, the assessments $S(e_i)$ ($i = 1, \cdots, L$), can be seen as $L$ pieces of evidence with weights $w_i$ ($i = 1, \cdots L$), in the frame of discernment $\mathcal{H} = \left\{ {{H_1}, \cdots ,{H_n}, \cdots ,{H_N}} \right\}$. Then each piece of evidence $S(e_i)$ can be characterized by a BBA ${m_i}(\cdot)$ as
\begin{equation}
{m_i}(A) = \left\{
\begin{array}{*{20}{l}}
\beta_{n,i}, &\text{~if~} A = \{H_n\} \\
1 - \sum\nolimits_{n = 1}^N {\beta_{n,i}}, &\text{~if~}A = \mathcal{H}\\
0, &\text{~others}.
\end{array} \right. \label{eq_BD}
\end{equation}
The quantity ${m_i}(\{H_n\} )$ represents the belief degree that supports the hypothesis that $e_i$ is assessed to the evaluation grade $H_n$, while ${m_i}(\mathcal{H})$ is the remaining belief degree unassigned to any individual grade after all evaluation grades have been considered for assessing ${e_i}$. As such, with $L$ basic attributes ${e_i}$, we obtain $L$ corresponding BBAs ${m_i}( \cdot )$ as quantified beliefs of the assessments for basic attributes. The problem now is how to generate an assessment for $y$, i.e., $S(y)$, represented by a BBA $m(\cdot)$, from ${m_i}(\cdot)$ and ${w_i}$, $i = 1, \cdots L$. Apparently this is a typical fusion problem in DST, and the usual way to handle this problem is to aggregate the $L$ BBAs using the discounting and combination operations.

In the about problem, the weights ${w_i}$ of basic attributes ${e_i}$ may be caused by different factors.
\begin{itemize}
  \item \emph{Reliability of attribute.} The expert may have different knowledge levels for different basic attributes. E.g., in evaluating the \emph{brakes} of a motorcycle, the expert is familiar with \emph{stopping power}, expert in \emph{braking} but has a low sensitivity for the \emph{feel at control} of a motorcycle.
  \item \emph{Importance of attribute.} The decision maker may have different preferences for different basics attributes. E.g., in evaluating the \emph{brakes} of a motorcycle, the decision maker believes \emph{stopping power} is more crucial than \emph{braking stability} and \emph{feel at control}.
\end{itemize}

In short, the weights ${w_i}$ of basic attributes can be interpreted either as the evaluation reliability or the relative importance of different basic attributes. In the next two subsections, we address the uncertain MADA problem considering the weights ${w_i}$ as reliability and importance of evidence, respectively. Interestingly, we find that the original ER algorithm follows the reliability discounting $\&$ combination scheme while the modified one follows the importance discounting $\&$ combination scheme.

\subsection{Analysis of the Original ER Algorithm in Terms of Reliability of Evidence}
If we consider the weights ${w_i}$ as the reliability of different basic attributes, the reliability discounting $\&$ combination scheme given in Section \ref{sec3B} is used to solve the uncertain MADA problem.

\begin{thm} \label{thm_OER}
The original ER algorithm, displayed as Eqs.(\ref{eq_OER_w})-(\ref{eq_OER_norm}), follows the reliability discounting $\&$ combination scheme displayed as Eqs.(\ref{eq_RD})-(\ref{eq_RC}).
\end{thm}

\begin{IEEEproof}
Firstly, with the reliability factors ${w_i}$ ($i = 1, \cdots, L$), Shafer's discounting operation displayed as Eq.(\ref{eq_RD}) is used for the $L$ BBAs ${m_i}( \cdot )$ to get the discounted BBAs $m_i^{{w_i}}( \cdot )$
\begin{equation}
m_i^{{w_i}}(A) = {w_i} \otimes {m_i}(A) = \left\{
\begin{array}{*{20}{l}}
{w_i}{\beta_{n,i}}, &\text{~if~} A = \{H_n\} \\
{w_i}\left(1 - \sum\nolimits_{n = 1}^N {\beta _{n,i}}\right) + (1 - {w_i}) = 1 - {w_i}\sum\nolimits_{n = 1}^N {\beta_{n,i}}, &\text{~if~} A = \mathcal{H}\\
0, &\text{~others}.
\end{array} \right.
\end{equation}
We can see that the weighted belief distribution displayed as Eq.(\ref{eq_OER_w}) in the original ER algorithm mathematically equals to the above formula. In other words, despite the weighted belief distribution in the original ER algorithm is formulated in a different way, it follows the reliability discounting operation in nature.

Then, we can use Dempster's rule of combination displayed as Eq.(\ref{eq_RC}) to combine the above $L$ reliability discounted BBAs $m_i^{{w_i}}( \cdot )$ to generate the BBA for the assessment of $y$. As we know that the Dempster's rule of combination is both commutative and associative, to reduce the computing complexity, the $L$ reliability discounted BBAs $m_i^{{w_i}}( \cdot )$ are combined recursively. Let $m_{I(i)}^D( \cdot )$ denote the combination result of the first $i$ reliability discounted BBAs, then it can be combined with the $(i + 1)$th reliability discounted BBA $m_{i + 1}^{{w_{i + 1}}}( \cdot )$ using Dempster's rule of combination as
\begin{equation}
m_{I(i + 1)}^D(A) = m_{I(i)}^D \oplus m_{i + 1}^{{w_{i + 1}}}(A) = \left\{
\begin{array}{*{20}{l}}
0, &\text{~for~} A = \emptyset\\
\frac{{\sum\limits_{B,C \in {2^\mathcal{H}};B \cap C = A} {m_{I(i)}^D(B)m_{i + 1}^{{w_{i + 1}}}(C)} }}{{1 - \sum\limits_{B,C \in {2^\mathcal{H}};B \cap C = \emptyset } {m_{I(i)}^D(B)m_{i + 1}^{{w_{i + 1}}}(C)}}}, &\text{~for~} A \in {2^\mathcal{H}} {\text{~and~}}A \ne \emptyset.
\end{array} \right.
\end{equation}
Because for all discounted BBAs $m_i^{{w_i}}( \cdot )$ ($i = 1, \cdots, L$), the focal elements are either singletons ${H_n}$ ($n = 1, \cdots, N$), or the ignorance set $\mathcal{H}$, so the focal elements of combination result $m_{I(i+1)}^D( \cdot )$ are also either singletons or the ignorance set $\mathcal{H}$. In this case, the above equation can be further simplified as
\begin{equation}
\begin{array}{*{20}{l}}
m_{I(i + 1)}^D(\{H_n\}) &= K_{I(i + 1)}\left[m_{I(i)}^D(\{H_n\})m_{i + 1}^{{w_{i + 1}}}(\{H_n\}) + m_{I(i)}^D(\{H_n\})m_{i + 1}^{{w_{i + 1}}}(\mathcal{H}) + m_{I(i)}^D(\mathcal{H})m_{i + 1}^{{w_{i + 1}}}(\{H_n\})\right], \\
&\text{~~~for~} n = 1,2, \cdots, N \\
m_{I(i + 1)}^D(\mathcal{H}) &= K_{I(i + 1)}\left[m_{I(i)}^D(\mathcal{H})m_{i + 1}^{{w_{i + 1}}}(\mathcal{H})\right] \\
K_{I(i + 1)} &= \left[1 - \sum\limits_{j = 1}^N {\sum\limits_{p = 1; p \ne j}^N {m_{I(i)}^D(\{H_j\})m_{{i + 1}}^{{w_{i + 1}}}(\{H_p\})}}\right]^{-1}, \text{~for~} i = 1,2, \cdots ,L - 1.
\end{array} \label{eq_Rresult}
\end{equation}
Comparing Eq.(\ref{eq_OER_agr}) with Eq.(\ref{eq_Rresult}), we can find that the aggregation process in the original ER algorithm and the Dempster's rule of combination lead essentially to mathematically equivalent formulations. Thus, after all the $L$ discounted BBAs are combined, we can get the BBA $m_{I(L)}^{E}( \cdot )$  for the assessment of $y$ and at last we get
\begin{equation}
\begin{array}{*{20}{l}}
\{{H_n}\} : &{\beta_n} &= m_{I(L)}^D(\{H_n\}), \text{~for~} n = 1,2, \cdots N\\
\mathcal{H} : &{\beta_\mathcal{H}} &= m_{I(L)}^D(\mathcal{H}).
\end{array}
\end{equation}

By now, it can be seen that the original ER algorithm reviewed in Section \ref{sec2C} formally follows the reliability discounting $\&$ combination scheme given in Section \ref{sec3B}.
\end{IEEEproof}

\subsection{Analysis of the Modified ER Algorithm in Terms of Importance of Evidence}
If we consider the weights ${w_i}$ as the importance of different basic attributes, the importance discounting $\&$ combination scheme developed in Section \ref{sec3C} is used to solve the uncertain MADA problem.

\begin{thm} \label{thm_MER}
The modified ER algorithm, displayed as Eqs.(\ref{eq_MER_w})-(\ref{eq_MER_norm}), follows the importance discounting $\&$ combination scheme displayed as Eqs.(\ref{eq_ID})-(\ref{eq_Norm}).
\end{thm}

\begin{IEEEproof}
Firstly, before using the importance discounting operation, as explained in Remark \ref{rek_ID}, the weights of basic attributes should be normalized such that $0 \le {w_i} \le 1$ and $\sum\nolimits_{i = 1}^L {{w_i}}  = 1$. This procedure is in agreement with the first step of the modified ER algorithm.

Then, with the normalized importance factors ${w_i}$ ($i = 1, \cdots, L$), the importance discounting operation displayed as Eq.(\ref{eq_ID}) is used for the $L$ BBAs ${m_i}( \cdot )$ to get the discounted IBBAs $m_i^{{w_i}}( \cdot )$
\begin{equation}
m_i^{{w_i}}(A) = {w_i} \odot {m_i}(A) = \left\{
\begin{array}{*{20}{l}}
{w_i}{\beta_{n,i}}, &\text{~if~} A = \{H_n\} \\
{w_i}\left(1 - \sum\nolimits_{n = 1}^N {\beta _{n,i}} \right), &\text{~if~} A = \mathcal{H}\\
1 - {w_i}, &\text{~if~} A = \Omega.
\end{array} \right.
\end{equation}
It can be seen that the decomposed probability masses ${\tilde m_{H,i}}$ and ${\bar m_{H,i}}$ displayed as Eq.(\ref{eq_MER_decom}) in the original ER algorithm equal $m_i^{{w_i}}(\mathcal{H})$ and $m_i^{{w_i}}(\Omega )$ respectively, though they are formulated in different ways.

After that, we can use the extend Dempster's rule of combination displayed as Eq.(\ref{eq_IC}) to combine the $L$ importance discounted IBBAs $m_i^{{w_i}}( \cdot )$ to generate the IBBA for the assessment of $y$. As pointed previously that the extended Dempster's rule of combination is also both commutative and associative, to reduce the computing complexity, the $L$ importance discounted IBBAs $m_i^{{w_i}}( \cdot )$ are combined recursively. Let $m_{I(i)}^{ED}( \cdot )$ denote the combination result of the first $i$ IBBAs, then it can be combined with the $(i + 1)$th IBBA $m_{i + 1}^{{w_{i + 1}}}( \cdot )$ using extended Dempster's rule of combination as
\begin{equation}
m_{I(i + 1)}^{ED}(A) = m_{I(i)}^{ED}\oplus m_{i + 1}^{{w_{i + 1}}}(A) = \left\{
\begin{array}{*{20}{l}}
0, &\text{~for~} A = \emptyset \\
\frac{{\sum\limits_{B,C \in {2^{{\mathcal{H}^ + }}};B \cap C = A} {m_{I(i)}^{ED}(B)m_{i + 1}^{{w_{i + 1}}}(C)} }}{{1 - \sum\limits_{B,C \in {2^{{\mathcal{H}^ + }}};B \cap C = \emptyset } {m_{I(i)}^{ED}(B)m_{i + 1}^{{w_{i + 1}}}(C)} }}, &\text{~for~} A \in {2^{{\mathcal{H}^ + }}} \text{~and~} A \ne \emptyset.
\end{array} \right.
\end{equation}
Because for all IBBAs $m_i^{{w_i}}( \cdot )$ ($i = 1, \cdots L$), the focal elements are either singletons ${H_n}$ ($n = 1, \cdots, N$), or the ignorance set $\mathcal{H}$ or the indecisiveness set $\Omega $, so the focal elements of combination result $m_{I(i+1)}^{ED}( \cdot )$ are also either singletons or the ignorance set $H$ or the indecisiveness set $\Omega $. In this case, the above equation can be further simplified as
\begin{equation}
\begin{array}{*{20}{l}}
m_{I(i + 1)}^{ED}(\{H_n\}) &= {K_{I(i + 1)}}\left[m_{I(i)}^{ED}(\{H_n\})m_{i + 1}^{{w_{i + 1}}}(\{H_n\}) + m_{I(i)}^{ED}(\{H_n\})(m_{i + 1}^{{w_{i + 1}}}(\mathcal{H}) + m_{i + 1}^{{w_{i + 1}}}(\Omega ))\right.\\
&\left.\text{~~} + (m_{I(i)}^{ED}(\mathcal{H}) + m_{I(i)}^{ED}(\Omega ))m_{i + 1}^{{w_{i + 1}}}(\{H_n\})\right], \text{~for~} n = 1,2, \cdots, N\\
m_{I(i + 1)}^{ED}(\mathcal{H}) &= {K_{I(i + 1)}}\left[m_{I(i)}^{ED}(\mathcal{H})m_{i + 1}^{{w_{i + 1}}}(\mathcal{H}) + m_{I(i)}^{ED}(\mathcal{H})m_{i + 1}^{{w_{i + 1}}}(\Omega ) + m_{I(i)}^{ED}(\Omega )m_{i + 1}^{{w_{i + 1}}}(\mathcal{H})\right]\\
m_{I(i + 1)}^{ED}(\Omega ) &= {K_{I(i + 1)}}\left[m_{I(i)}^{ED}(\Omega )m_{i + 1}^{{w_{i + 1}}}(\Omega )\right]\\
{K_{I(i + 1)}} &= {\left[1 - \sum\limits_{j = 1}^N {\sum\limits_{p = 1; p \ne j}^N {m_{I(i)}^{ED}(\{H_j\})m_{i + 1}^{{w_{i + 1}}}(\{H_p\})} }\right]^{ - 1}}, \text{~for~}i = 1,2, \cdots ,L - 1.
\end{array} \label{eq_Iresult}
\end{equation}
Comparing Eq.(\ref{eq_MER_agr}) with Eq.(\ref{eq_Iresult}), we can find that the aggregation process with decomposed masses in the modified ER algorithm and the extended Dempster's rule of combination lead essentially to mathematically equivalent formulations. Thus, after all the $L$ IBBAs are combined, we can get the IBBA $m_{I(L)}^{ED}( \cdot )$ for the assessment of $y$.

At last, the combined IBBA $m_{I(L)}^{ED}( \cdot )$ will be normalized to normal BBA for decision making support with Eq.(\ref{eq_Norm}) as
\begin{equation}
\begin{array}{*{20}{l}}
\{{H_n}\} : &{\beta _n} &= \frac{{m_{I(L)}^{ED}(\{H_n\})}}{{1 - m_{I(L)}^{ED}(\Omega )}}, \text{~for~} n = 1,2, \cdots, N\\
\mathcal{H} : &{\beta _\mathcal{H}} &= \frac{{m_{I(L)}^{ED}(\mathcal{H})}}{{1 - m_{I(L)}^{ED}(\Omega )}}.
\end{array}
\end{equation}
This process is in agreement with the last step of the modified ER algorithm though they are explained in different ways.

By now, it can be seen that the modified ER algorithm reviewed in Section \ref{sec2C} formally follows the importance discounting $\&$ combination scheme given in Section \ref{sec3C}.
\end{IEEEproof}

\begin{rmk}
Theorem \ref{thm_OER} and Theorem \ref{thm_MER} reveal the potential nature of the original and modified ER algorithms in DST framework. For the two ER algorithms, we cannot simply say which one is right or more rational, because they are just two different algorithms focusing on different uncertain MADA problems (i.e. the reliability of attribute is considered in the original ER algorithm, while the importance of attribute is considered in the modified ER algorithm). These findings can help us to select an appropriate ER algorithm for a specific MADA problem.
\end{rmk}

\subsection{Discussion of the Synthesis Axioms}
As summarized in Section \ref{sec2C}, intending to develop theoretically sound methods for dealing with uncertain MADA problems, Yang and Xu \cite{Yang02b} proposed four synthesis axioms, i.e. \emph{Independence}, \emph{Consensus}, \emph{Completeness} and \emph{Incompleteness},  within the evidential reasoning assessment framework. They believe that a rational aggregation process needs to satisfy these synthesis axioms and so modified the original ER algorithm because some of the synthesis axioms were not satisfied. However, the authors failed to give a reasonable explanation for the rationality of the synthesis axioms. The problem is whether only the algorithms satisfying the four axioms are rational while all others are irrational. As we have revealed the potential nature of the original and modified ER algorithms in DST framework, it provides a new way to explain the synthesis axioms. In the following, we will discuss the synthesis axioms from the aspects of reliability and importance of evidence.

\begin{figure}[!ht]
\centering
\includegraphics[scale=0.7]{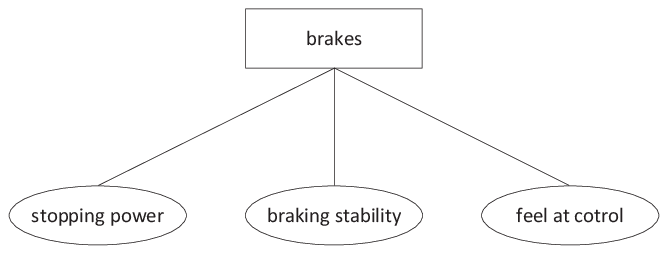}
\caption{Evaluation hierarchy for brakes performance assessment}
\label{fig3}
\end{figure}

In order to explain the synthesis axioms in a more comprehensible way, we consider the following simple MADA example.

\emph{Example 1} (Brakes performance assessment). As in Fig.\ref{fig3}, we study the problem of evaluating the performance of brakes of motorcycles considering three basic attributes, namely \emph{stopping power}, \emph{braking stability} and \emph{feel at control}. The evaluation grades in Eq.(\ref{eq_evaluation_grade}) is used here to evaluate the basic attributes. For this problem, the four synthesis axioms can be described as follows.
\begin{itemize}
  \item If no basic attribute is assessed to \emph{excellent} grade at all, then the brakes should not be assessed to \emph{excellent} grade either. (\emph{Independence})
  \item If all basic attributes are precisely assessed to \emph{excellent} grade, then the brakes should also be precisely assessed to \emph{excellent} grade. (\emph{Consensus})
  \item If all basic attributes are completely assessed to a subset of grades, e.g., $\{good, excellent\}$, then the brakes should be completely assessed to $\{good, excellent\}$ as well. (\emph{Completeness})
  \item If any basic assessment is incomplete, e.g., the \emph{stopping power} is not completely assessed to \emph{excellent} grade, then the brakes should not be completely assessed to \emph{excellent} grade. (\emph{Incompleteness})
\end{itemize}

As demonstrated in \cite{Yang02b}, the original ER algorithm fails to satisfy the latter three axioms. As we have proved that the original ER algorithm actually follows the reliability discounting $\&$ combination scheme, so we can explain the reasons from the viewpoint of reliability of basic attributes. Here, we take the \emph{consensus} axiom for explanation. Suppose all basic attributes are precisely assessed to \emph{excellent} grade. But, e.g., considering the expert has a low sensitivity for the \emph{feel at control} of brakes, wrong assessment may be provided (maybe the \emph{feel at control} is \emph{poor} in reality), and so the brakes should not be precisely assessed to \emph{excellent} grade. Similar reasons can be given for \emph{completeness} and \emph{incompleteness} axioms. In brief, the latter three axioms are no longer reasonable when reliability of basic attributes is considered in the assessment.

While, the modified ER algorithm developed in \cite{Yang02b} satisfies all the four axioms. As we have proved that the modified ER algorithm actually follows the importance discounting $\&$ combination scheme, so we can explain the reasons from the viewpoint of importance of basic attributes. Here, we also take the \emph{consensus} axiom for explanation. Suppose all basic attributes are precisely assessed to \emph{excellent} grade. As the importance reflects the subjective preferences for different basic attributes and it does no matter with their reliability, so the importance discounting only assigns weight to each basic attribute rather than changes their assessments, and thus it's intuitively reasonable that the brakes should also be precisely assessed to \emph{excellent} grade. The satisfaction for the other axioms are due to similar reasons. In brief, when importance of basic attributes is considered in the assessment, the four synthesis axioms are still applicable.

Through the above analysis, it can be seen that the four synthesis axioms proposed by Yang and Xu \cite{Yang02b} are only reasonable when the importance of basic attributes is considered and when we take the reliability of basic attributes into consideration, they are no longer applicable. So these synthesis axioms cannot be used as criteria to check the rationality of one attribute aggregation algorithm.

\begin{rmk}
With the above conclusion, we can explain the result of Durbach's empirical test \cite{Durbach12} mentioned in Introduction more reasonably. Actually, in his design of empirical test for the four synthesis, the reliability of basic attributes is considered under which circumstance the four synthesis axioms are no longer rational.
\end{rmk}

\section{Extension of the ER Algorithm}  \label{sec5}
As analyzed in the preceding section, in uncertain MADA problems, the weights of basic attributes can be interpreted either as the reliability or the importance of different basic attributes. We find that the original ER algorithm only considers the reliability of basic attributes, whereas the modified ER algorithm only considers the importance. However, in many real-world MADA problems, the reliability and importance of basic attributes may coexist with each other and so we need to take into account both of them to make a more rational decision. In this section, a general reliability-importance discounting $\&$ combination scheme is given in DST framework. Then, we use this scheme to address the uncertain MADA problem and develop an extended ER algorithm which provides a more general attribute aggregation scheme.

\subsection{Reliability-Importance Discounting $\&$ Combination Scheme}
In this subsection, we focus on the combination of sources of evidence considering both their reliability and importance. For each source of evidence characterized by its BBA $m( \cdot )$, as the reliability and importance are independent with each other, we can discount the original BBA $m( \cdot )$ by $\alpha $ with reliability discounting operation and then discount the result ${m^\alpha }( \cdot )$ by $\beta $ with importance discounting operation.

\begin{defi}[Reliability-importance discounting operation]
The reliability-importance discounting operation of a source of evidence with the reliability factor $\alpha  \in [0,1]$, the importance factor $\beta  \in [0,1]$ and the associated BBA $m( \cdot )$ in the frame of discernment $\Theta $ is
\begin{equation}
\beta \odot \left( {\alpha  \otimes m(A)} \right) \buildrel \Delta \over = {m^{\alpha ,\beta }}(A) = \left\{
\begin{array}{*{20}{l}}
\alpha \beta m(A), &\text{~for~} A \in {2^\Theta}, \text{~} A \ne \Theta \\
\alpha \beta m(\Theta) + (1 - \alpha )\beta, &\text{~for~} A = \Theta \\
1 - \beta, &\text{~for~}A = \Omega.
\end{array} \right. \label{eq_RID}
\end{equation}
where, $\Omega $ has the same meaning as in Eq.(\ref{eq_ID}).
\end{defi}

Because the result of reliability-importance discounting operation is an IBBA, after all the involved evidences are discounted with Eq.(\ref{eq_RID}), the extended Dempster's rule of combination displayed as Eq.(\ref{eq_IC}) can be used to get the combined result.

The above reliability-importance discounting $\&$ combination scheme provides a general scheme for combining sources of evidence to take into account both the reliability and importance. It can be seen that, when ${\beta _i} = 1, \text{~} i = 1, \cdots, L$ (the sources have full importance), the reliability-importance discounting $\&$ combination scheme simplifies to the reliability discounting $\&$ combination scheme presented in Section \ref{sec3B}; when ${\alpha _i} = 1, \text{~} i = 1, \cdots, L$ (the sources have full reliability), the importance discounting $\&$ combination scheme presented in Section \ref{sec3C} is obtained; and when both ${\alpha _i} = 1$ and ${\beta _i} = 1$, $i = 1, \cdots, L$ (the sources have both full importance and reliability), the original BBA ${\text m} ( \cdot )$ keeps unchanged and the classical Dempster's rule of combination is used to integrate the sources of evidence.

\subsection{Extended ER (E$^2$R) Algorithm for MADA with Uncertainty}
We also consider the assessment problem with two-level evaluation hierarchy, as shown in Fig.\ref{fig2}, but the difference is each basic attribute ${e_i}$ has both reliability ${\alpha _i} \in [0,1]$ and importance ${\beta _i} \in [0,1]$, $i = 1, \cdots L$. As explained in Remark \ref{rek_ID}, the importance factor ${\beta _i}$ should be normalized such that $0 \le {\beta_i} \le 1$ and $\sum\nolimits_{i = 1}^L {{\beta_i}}  = 1$. Now, we develop the extended ER (E$^2$R) algorithm for MADA as follows.

\textbf{E$^2$R algorithm:}
\begin{enumerate}
  \item \emph{Reliability-importance discounting}. The $L$ BBAs ${m_i}( \cdot )$ formulated as Eq.(\ref{eq_BD}) are discounted with the reliability-importance discounting operation displayed as Eq.(\ref{eq_RID})
\begin{equation}
{m_i}^{{\alpha_i},{\beta_i}}(A) = {\beta _i}{\odot}\left( {{\alpha _i} \otimes {m_i}(A)} \right) = \left\{
\begin{array}{*{20}{l}}
{\alpha_i}{\beta_i}{m_i}(A), &\text{~for~} A = \{H_n\}\\
{\alpha _i}{\beta _i}{m_i}(\mathcal{H}) + (1 - \alpha )\beta , &\text{~for~}A = \mathcal{H}\\
1 - {\beta_i}, &\text{~for~} A = \Omega.
\end{array} \right.
\end{equation}

  \item \emph{Aggregation process}. Combine the $L$ discounted IBBAs $m_i^{{\alpha _i},{\beta _i}}( \cdot )$ with the extend Dempster's rule of combination displayed as Eq.(\ref{eq_IC}) recursively to generate the IBBA for the assessment of $y$
\begin{equation}
m_{I(i + 1)}^{ED}(A) = m_{I(i)}^{ED} \oplus m_{i + 1}^{{\alpha _{i + 1}},{\beta _{i + 1}}}(A) = \left\{
\begin{array}{*{20}{l}}
0, &\text{~for~} A = \emptyset \\
\frac{{\sum\limits_{B,C \in {2^{{\mathcal{H}^ + }}};B \cap C = A} {m_{I(i)}^{ED}(B)m_{i + 1}^{{\alpha _{i + 1}},{\beta _{i + 1}}}(C)} }}{{1 - \sum\limits_{B,C \in {2^{{\mathcal{H}^ + }}};B \cap C = \emptyset } {m_{I(i)}^{ED}(B)m_{i + 1}^{{\alpha _{i + 1}},{\beta _{i + 1}}}(C)} }}, &\text{~for~} A \in {2^{{\mathcal{H}^ + }}} {\text{~and~}} A \ne \emptyset.
\end{array} \right.
\end{equation}
To this special problem, because for all discounted IBBAs $m_i^{{\alpha _i},{\beta _i}}( \cdot )$ ($i = 1, \cdots L$), the focal elements are either singletons $\{H_n\}$
($n = 1, \cdots, N$), or the set $\mathcal{H}$ or $\Omega$, so the above equation can be further simplified as
\begin{equation}
\begin{array}{*{20}{l}}
m_{I(i + 1)}^{ED}(\{H_n\}) &= {K_{I(i + 1)}}\left[m_{I(i)}^{ED}(\{H_n\})m_{i + 1}^{{\alpha _{i + 1}},{\beta _{i + 1}}}(\{H_n\}) + m_{I(i)}^{ED}(\{H_n\})(m_{i + 1}^{{\alpha _{i + 1}},{\beta _{i + 1}}}(\mathcal{H}) + m_{i + 1}^{{\alpha _{i + 1}},{\beta _{i + 1}}}(\Omega ))\right.\\
&\left.\text{~~} + (m_{I(i)}^{ED}(\mathcal{H}) + m_{I(i)}^{ED}(\Omega ))m_{i + 1}^{{\alpha _{i + 1}},{\beta _{i + 1}}}(\{H_n\})\right], \text{~for~} n = 1,2, \cdots, N\\
m_{I(i + 1)}^{ED}(\mathcal{H}) &= {K_{I(i + 1)}}\left[m_{I(i)}^{ED}(\mathcal{H})m_{i + 1}^{{\alpha _{i + 1}},{\beta _{i + 1}}}(\mathcal{H}) + m_{I(i)}^{ED}(\mathcal{H})m_{i + 1}^{{\alpha _{i + 1}},{\beta _{i + 1}}}(\Omega ) + m_{I(i)}^{ED}(\Omega )m_{i + 1}^{{\alpha _{i + 1}},{\beta _{i + 1}}}(\mathcal{H})\right]\\
m_{I(i + 1)}^{ED}(\Omega ) &= {K_{I(i + 1)}}\left[m_{I(i)}^{ED}(\Omega )m_{i + 1}^{{\alpha _{i + 1}},{\beta _{i + 1}}}(\Omega )\right]\\
{K_{I(i + 1)}} &= {\left[1 - \sum\limits_{j = 1}^N {\sum\limits_{p = 1; p \ne j}^N {m_{I(i)}^{ED}(\{H_j\})m_{i + 1}^{{\alpha _{i + 1}},{\beta _{i + 1}}}(\{H_p\})} }\right]^{ - 1}}, \text{~for~}i = 1,2, \cdots ,L - 1.
\end{array} \label{eq_RIresult}
\end{equation}

  \item \emph{Belief degree normalization}. After all the $L$ IBBAs are combined, normalize the fusion result $m_{I(L)}^{ED}( \cdot )$ by redistributing the mass of belief committed to the set $\Omega $ to the other focal elements proportionally
\begin{equation}
\begin{array}{*{20}{l}}
\{{H_n}\} : &{\beta _n} &= \frac{{m_{I(L)}^{ED}({H_n})}}{{1 - m_{I(L)}^{ED}(\Omega )}}, \text{~for~} n = 1,2, \cdots N\\
\mathcal{H} : &{\beta _H} &= \frac{{m_{I(L)}^{ED}(H)}}{{1 - m_{I(L)}^{ED}(\Omega )}}.
\end{array}
\end{equation}
\end{enumerate}

The above E$^2$R algorithm provides a general scheme for uncertain MADA problems by taking into account both the reliability and importance of basic attributes. The original ER algorithm is the special case that all basic attributes have same importance (${\beta _i} = 1, \text{~} i = 1, \cdots, L$) and the modified one is the special case that all basic attributes have full reliability (${\alpha _i} = 1, \text{~} i = 1, \cdots, L$).

\section{Numerical Study: Motorcycle Performance Assessment} \label{sec6}
\begin{figure}[!h]
\centering
\includegraphics[scale=0.8]{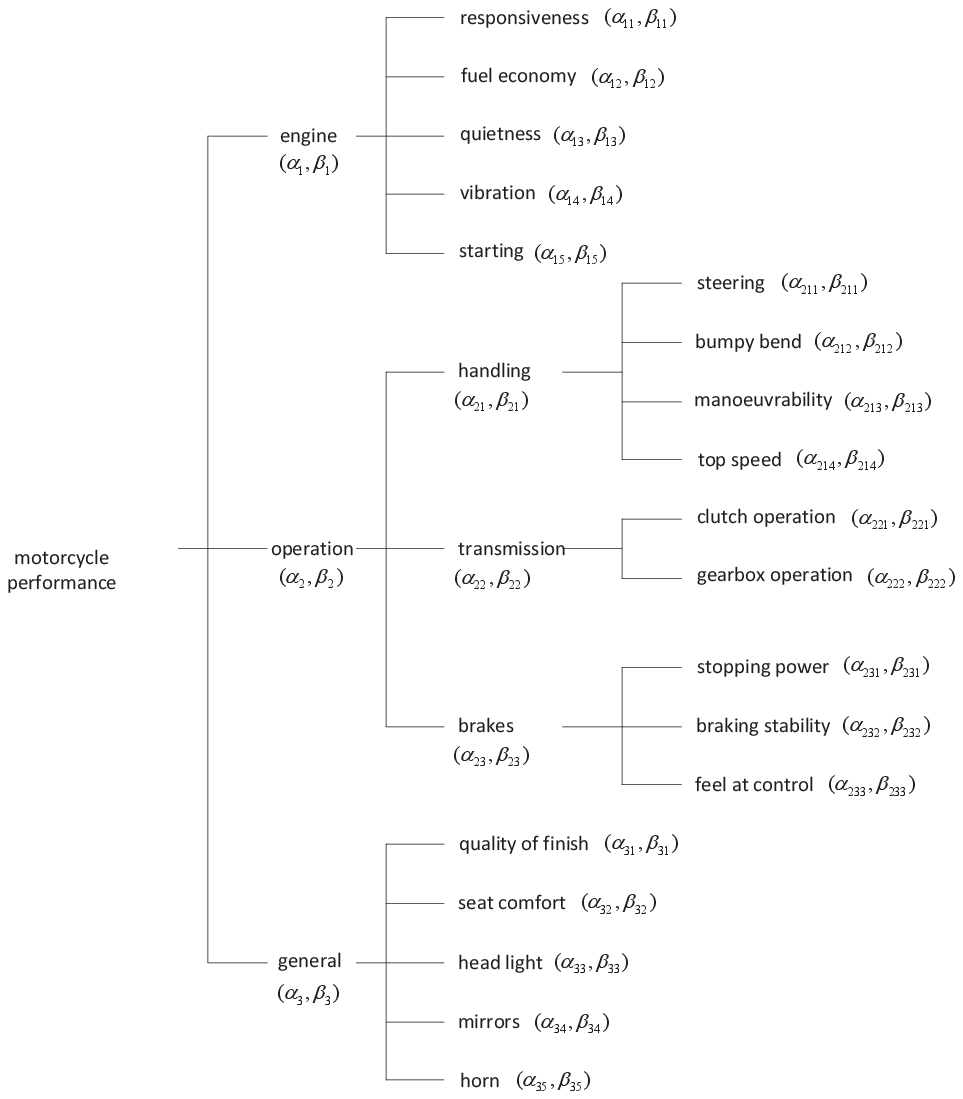}
\caption{Evaluation hierarchy for motorcycle performance assessment}
\label{fig4}
\end{figure}

In this section, we give a study for the problem of motorcycle performance assessment \cite{Isitt90,Yang02b}. In this study, the decision alternatives are composed of four types of motorcycles, namely, Kawasaki, Yamaha, Honda, and BMW. The performance of each motorcycle is evaluated based on three major attributes: \emph{quality of engine}, \emph{operation}, and \emph{general finish}. Because these attributes are too general to assess directly, they are decomposed into more detailed sub-attributes (until they could not be decomposed any longer, called basic attributes) to facilitate the assessment. At last, we can obtain an evaluation hierarchy as shown in Fig.\ref{fig4}, where the reliability and relative importance of attributes are defined by ${\alpha _i}$,  ${\alpha _{i,j}}$, ${\alpha _{i,j,k}}$ and ${\beta _i}$,  ${\beta _{i,j}}$, ${\beta _{i,j,k}}$ for the attributes at levels 1, 2, and 3, respectively.

Using the five-grade evaluation scale as given in Eq.(\ref{eq_evaluation_grade}), the decision matrix for this motorcycle performance assessment problem is given in TABLE \ref{tab1}, where P, I, A, G, and E are the abbreviations of \emph{poor}, \emph{indifferent}, \emph{average}, \emph{good}, and \emph{excellent}, respectively. The number after each grade denotes the belief degree to which a basic attribute is assessed to this grade, which is usually given by experts. For example, \textquotedblleft E(0.8)\textquotedblright~ means \textquotedblleft \emph{excellent} to a degree of 0.8\textquotedblright. Besides, the two numbers in braces following each attribute denote the evaluation reliability and the relative importance of this attribute, respectively. For example, \textquotedblleft fuel economy(0.8,0.1)\textquotedblright~ means \textquotedblleft the assessment for \emph{fuel economy} has a degree of 0.8 reliability and the attribute \emph{fuel economy} has a degree of 0.1 relative importance comparing with other attributes at the same level\textquotedblright. Usually, evaluation reliability of each attribute is related to the expert's knowledge level in this particular field. Because only the basic attributes are evaluated directly by the experts, so the evaluation reliability of the general attributes should be calculated based on the evaluation reliability of their sub-attributes (arithmetic average is used here for its convenience). But the relative importance of each attribute reflects the decision maker's subjective preference and it can be obtained through several weight assignment methods \cite{Xu04,Roberts02}.

\begin{table}[!ht]
\caption{Decision matrix for motorcycle performance assessment}
\begin{center}
\begin{tabular}{*{8}{|l}|} \hline
\multicolumn{3}{|c|}{\textbf{General attributes}} & \textbf{Basic attributes}  & \multicolumn{4}{|c|}{\textbf{Types of motorcycle(alternatives)}}\\ \cline{5-8}
\multicolumn{3}{|c|}{}   &                         & \textbf{Kawasaki}     & \textbf{Yamaha}      & \textbf{Honda}     & \textbf{BMW}     \\ \hline
 &\multicolumn{2}{|c|}{} &responsiveness(0.6,0.2)  & E(0.8)       & G(0.3),E(0.6) & G(1.0)        & I(1.0)  \\
 &\multicolumn{2}{|c|}{} &fuel economy(0.7,0.2)    & A(1.0)       & I(1.0)        & I(0.5),A(0.5) & E(1.0)  \\
 &\multicolumn{2}{|c|}{engine(0.7,0.4)} &quietness(0.4,0.1) & I(0.5),A(0.5)   & A(1.0)  & G(0.5),E(0.3) & E(1.0)  \\
 &\multicolumn{2}{|c|}{} &vibration(0.9,0.2)       & G(1.0)       & I(1.0)        & G(0.5),E(0.5) & P(1.0)  \\
 &\multicolumn{2}{|c|}{} &starting(0.9,0.3)        & G(1.0)       & A(0.6),G(0.3) & G(1.0)        & A(1.0)  \\ \cline{2-8}
 & &                     &steering(0.9,0.3)        & E(0.9)       & G(1.0)        & A(1.0)        & A(0.6)  \\
 & &handing              &bumpy bends(0.8,0.3)     & A(0.5),G(0.5)& G(1.0)        & G(0.8),E(0.1) & P(0.5),I(0.5)  \\
 & &(0.8,0.4)            &maneuverability(0.5,0.2) & A(1.0)       & E(0.9)        & I(1.0)        & P(1.0)  \\
 & &                     &top speed stability(1.0,0.2) & E(1.0)   & G(1.0)        & G(1.0)        & G(0.6),E(0.4)  \\ \cline{3-8}
motorcycle &operation &transmission &clutch operation(0.6,0.3)  & A(0.8)       & G(1.0)        & E(0.85)        & I(0.2),A(0.8)  \\
performance &(0.7,0.3) &(0.7,0.3)  &gearbox operation(0.8,0.7)  & A(0.5),G(0.5) & I(0.5),A(0.5)  & E(1.0)        & P(1.0)  \\ \cline{3-8}
 & &                     &stopping power(0.9,0.4)   & G(1.0)       & A(0.3),G(0.6) & G(0.6)        & E(1.0)  \\
 & &brakes               &braking stability(0.7,0.3)& G(0.5),E(0.5)& G(1.0)        & A(0.5),G(0.5) & E(1.0)  \\
 & &(0.6,0.3)                     &feel at control(0.2,0.3)  & P(1.0)       & G(0.5),E(0.5) & G(1.0)        & G(0.5),E(0.5)  \\ \cline{2-8}
 &\multicolumn{2}{|c|}{} &quality of finish(0.6,0.1)& P(0.5),I(0.5)& G(1.0)        & E(1.0)        & G(0.5),E(0.5)  \\
 &\multicolumn{2}{|c|}{} &seat comfort(0.9,0.2)     & G(1.0)       & G(0.5),E(0.5) & G(0.6)        & E(1.0)  \\
 &\multicolumn{2}{|c|}{general(0.8,0.3)} &headlight(0.9,0.4) & G(1.0)   & A(1.0)  & E(1.0)         & G(0.5),E(0.5)  \\
 &\multicolumn{2}{|c|}{} &mirrors(0.8,0.2)          & G(0.5),E(0.5)& G(0.5),E(0.5) & E(1.0)        & G(1.0)  \\
 &\multicolumn{2}{|c|}{} &horn(0.8,0.1)             & A(1.0)       & G(1.0)    & G(0.5),E(0.5)     & E(1.0)  \\ \hline
\end{tabular}
\end{center} \label{tab1}
\end{table}

For this motorcycle performance assessment problem, as both the evaluation reliability and relative importance of different attributes are involved, the more general extended ER (E$^2$R) algorithm proposed in this paper should be used. For the purpose of comparison, we also generate the aggregation results of the original ER (OER) algorithm and the modified ER (MER) algorithm. As we know that the OER algorithm follows the reliability discounting $\&$ combination scheme, so only the evaluation reliability of each attribute is considered, whereas the MER algorithm follows the importance discounting $\&$ combination scheme, so only the relative importance of each attribute is considered.

\begin{figure}[!ht]
\centering
\includegraphics[scale=0.8]{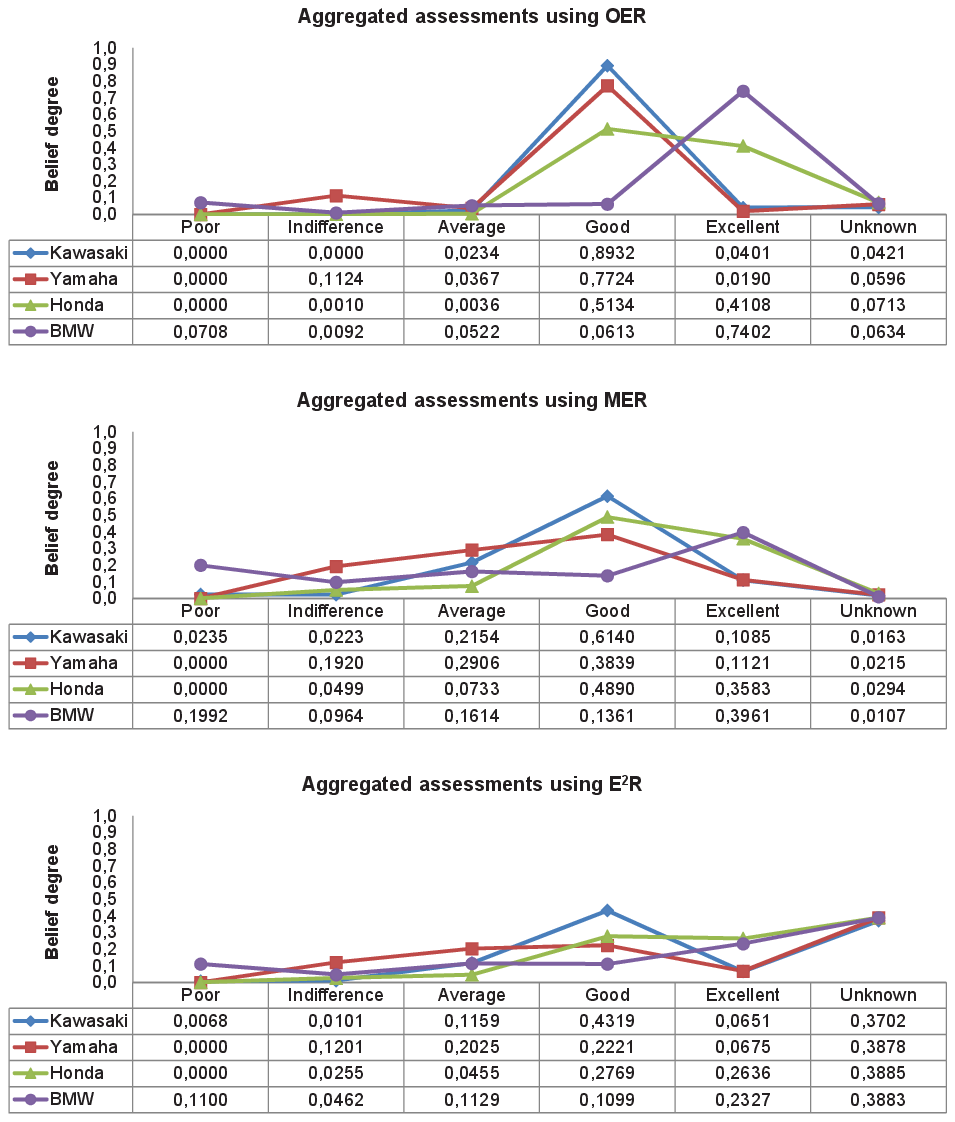}
\caption{Aggregated assessments with different algorithms}
\label{fig5}
\end{figure}

Fig.\ref{fig5} gives the aggregation results for the four types of motorcycles with three different algorithms. As different kinds of additional information are considered in the aggregation process for the three method, the aggregation results are quite different. We can see that more mass is assigned to \emph{Unknown} for the proposed E$^2$R algorithm, which is mainly because in this example, the evaluation reliability and relative importance of some attributes are in conflict with each other (i.e., the attribute with high reliability may have low importance or the attribute with low reliability may have high importance). Therefore, the the proposed E$^2$R algorithm can better model the global ignorance caused by conflict information.
~\\

Further, for the purpose of decision making, the pignistic probability as Eq.(\ref{eq_BetP}) is used here in order to assign the \emph{Unknown} mass ${\beta _\mathcal{H}}$ to the individual evaluation grades ${\beta _n}$
\begin{equation}
{\beta'_n} = {\beta _n} + \frac{1}{N}{\beta _\mathcal{H}}, \text{~for~} n = 1, \cdots ,N. \label{eq_assignP}
\end{equation}
Then, the expected utility of an alternative on the general attribute $y$ can be expressed as
\begin{equation}
u(y) = \sum\limits_{n = 1}^N {{{\beta '}_n}u({H_n})}, \label{eq_utility}
\end{equation}
where, $u( \cdot ):\mathcal{H} \to [0,1]$ is a utility function as define in \cite{Yang02b}. To this motorcycle performance assessment problem, we can set
\[u(P) = 0.2,\text{~} u(I) = 0.4,\text{~} u(A) = 0.6,\text{~} u(G) = 0.8,\text{~} u(E) = 1.\]

Using Eqs.(\ref{eq_assignP}) and (\ref{eq_utility}), we can obtain the expected utilities of the four types of motorcycles according to different algorithms as shown in TABLE \ref{tab2}. Further, the ranking orders based on the expected utilities are given in TABLE \ref{tab3}. Not surprisingly, the decision results are different for the three algorithms because different kinds of additional information are used in the decision process. It's believed that the E$^2$R algorithm gives the most reasonable decision result as both the evaluation reliability and the relative importance of the attributes are considered.

\begin{table}[!ht]
\caption{Expected utilities of four types of motorcycles}
\begin{center}
\begin{tabular}{*{5}{|l}|} \hline
\textbf{Algorithm}          &\textbf{Kawasaki}	&\textbf{Yamaha}	   &\textbf{Honda}	  &\textbf{BMW}      \\ \hline
\textbf{OER algorithm}	    &0.7943	    &0.7396	   &0.8668	  &0.8782   \\ \hline
\textbf{MER algorithm}	    &0.7347	    &0.6607	   &0.8073	  &0.7377   \\ \hline
\textbf{E$^2$R algorithm}	&0.7077	    &0.6474	   &0.7557	  &0.6618   \\ \hline
\end{tabular}
\end{center} \label{tab2}
\end{table}

\begin{table}[!ht]
\caption{Ranking orders of four types of motorcycles}
\begin{center}
\begin{tabular}{*{2}{|l}|} \hline
\textbf{Algorithm}	        &\textbf{Ranking order}              \\ \hline
\textbf{OER algorithm}	    &BMW   $\succ$ Honda    $\succ$ Kawasaki  $\succ$ Yamaha      \\ \hline
\textbf{MER algorithm}	    &Honda $\succ$ BMW      $\succ$ Kawasaki  $\succ$ Yamaha      \\ \hline
\textbf{E$^2$R algorithm}	&Honda $\succ$ Kawasaki $\succ$ BMW       $\succ$ Yamaha      \\ \hline
\end{tabular}
\end{center} \label{tab3}
\end{table}

\section{Concluding Remarks} \label{sec7}
In this paper, the original and modified ER algorithms for MADA with uncertainty are analyzed in DST framework, and we find that the original ER algorithm actually follows the reliability discounting $\&$ combination scheme, while the modified ER algorithm follows the importance discounting $\&$ combination scheme. These findings reveal the potential nature of these algorithms and provide us a guide to select the appropriate ER algorithm for a specific MADA problem. Further, we explain the synthesis axioms from the aspects of reliability and importance of evidence and reveal that the four synthesis axioms are only reasonable when the importance of basic attributes is considered. Finally, we introduce an extension of the ER algorithm, called E$^2$R algorithm, which provides a more general attribute aggregation scheme for MADA with uncertainty.
The capability of the proposed E$^2$R algorithm for addressing both the evaluation reliability and relative importance of different attributes is demonstrated through a motorcycle performance assessment problem.

\ifCLASSOPTIONcaptionsoff
  \newpage
\fi



%

\bibliographystyle{IEEEtran}
\bibliography{MyBib}

%







\end{document}